\documentclass{article}

 \usepackage[main, final]{neurips_2026}

\usepackage[utf8]{inputenc} 
\usepackage[T1]{fontenc}    
\usepackage{hyperref}       
\usepackage{url}            
\usepackage{booktabs}       
\usepackage{amsfonts}       
\usepackage{nicefrac}       
\usepackage{microtype}      
\usepackage{xcolor}         
\usepackage[textsize=tiny]{todonotes}
\usepackage{amsmath}

\newcommand{\CRrevisioncolor}{black}
\DeclareRobustCommand{\CRchange}[1]{{\color{\CRrevisioncolor}#1}}
\newcommand{\rev}[1]{{\color{black}{#1}}}

\title{
Neurosymbolic Learning \\ for Inference-Time Argumentation}

\author{%
  \textbf{Gabriel Freedman}\thanks{Equal contribution.} \quad
  \textbf{Adam Dejl}\footnotemark[1] \quad
  \textbf{Adam Gould}\footnotemark[1] \quad
  \textbf{Mansi}\footnotemark[1] \\
  \textbf{Lihu Chen} \quad
  \textbf{Junqi Jiang} \quad
  \textbf{Francesca Toni} \\
 Department of Computing, Imperial College London\\
  \texttt{\{g.freedman22, ad5518, amg2118, m24\}@imperial.ac.uk}\\
    \texttt{\{junqi.jiang, lihu.chen, ft\}@imperial.ac.uk}
}

\begin{document}

\newcommand{\complete}[2]{\todo[linecolor=orange,backgroundcolor=orange!25,bordercolor=orange]{#1: \textbf{Complete:} #2}}
\newcommand{\tocite}[2]{\todo[linecolor=pink,backgroundcolor=pink!25,bordercolor=pink]{#1: \textbf{To Cite:} #2}}
\newcommand{\unsure}[2]{\todo[linecolor=blue,backgroundcolor=blue!5,bordercolor=blue]{#1: \textbf{Unsure:} #2}}
\newcommand{\change}[2]{\todo[linecolor=red,backgroundcolor=red!25,bordercolor=red]{#1: \textbf{Change:} #2}}
\newcommand{\info}[2]{\todo[linecolor=olive,backgroundcolor=olive!25,bordercolor=olive]{#1: \textbf{Info:} #2}}
\newcommand{\improvement}[2]{\todo[linecolor=purple,backgroundcolor=purple!25,bordercolor=purple]{#1: \textbf{Improve:} #2}}
\newcommand{\cut}[2]{\todo[linecolor=yellow,backgroundcolor=yellow!25,bordercolor=yellow]{#1: \textbf{Potential Cut:} #2}}
\newcommand\todoin[2]{\todo[inline, caption={Outline}, linecolor=green,backgroundcolor=green!25,bordercolor=green,size=small]{\begin{minipage}{\textwidth-4pt} #1 \textbf{Outline:} #2\end{minipage}}}

\newcommand{\itemtodo}[0]{\item[$\Box$]}
\newcommand{\itemdone}[0]{\item[$\checkmark\hspace{-0.85em}\Box$]}
\newcommand{\itemdoing}[0]{\item[\text{\large$\circ$}$\hspace{-0.69em}\Box$]}

\newcommand{\itemcancel}[0]{\item[$\times\hspace{-0.75em}\Box$]}

\newcommand{\itemmaybe}[0]{\item[$?\hspace{-0.6em}\Box$]}

\newcommand{\basescoreparam}[0]{\phi}
\newcommand{\basescoreparamllm}[0]{\basescoreparam_{llm}}
\newcommand{\basescoreparamhead}[0]{\basescoreparam_{reg}}
\newcommand{\basescorefunc}[0]{f_\basescoreparam}
\newcommand{\basescorellm}[0]{f_{\basescoreparamllm}}
\newcommand{\basescorehead}[0]{f_{\basescoreparamhead}}

\maketitle

\begin{abstract}

Claim verification is an important problem in 
high-stakes settings, including health 
and finance
. When information
underpinning claims is incomplete or conflicting,
uncertain answers may be more appropriate than binary true or false classifications. In all cases, faithful explanations of the considerations determining the final verdict are crucial.
We
introduce 
 \emph{inference-time argumentation (ITA)}, a trainable neurosymbolic framework for ternary claim verification in which a
  formal argumentation semantics giving the strength of claims
  and arguments
  is
  used both (i) to guide LLM training as models learn to \emph{generate arguments}
  for and against claims
  and \emph{assign them base scores} (representing intrinsic strengths) and (ii) to compute ternary (true/false/uncertain) predictions from generated, scored arguments.
As a result, at training time, argument generation and scoring can be optimised  according to the quality of the induced argumentative predictions. 
Moreover, at inference time,
the final prediction is faithful, by construction, to the arguments and scores determining the verdict, rather than being justified by a potentially unfaithful post-hoc reasoning trace as in conventional reasoning models. 
We finally show that, on two 
datasets for ternary claim verification, ITA improves upon argumentative baselines and can perform competitively against non-argumentative direct-prediction baselines, while providing verdicts that are computed deterministically from explicit, inspectable argumentative structures.
\CRchange{Our code, models and data are available at \url{https://github.com/CLArg-group/inference-time-argumentation}.}
\end{abstract}

\section{Introduction}
Claim verification
amounts to ascertaining the veracity of textual claims
.
It is a widely studied problem, recently as a potential application for LLMs~\citep{claimverification-survey}, and is used in many high-stakes settings, including healthcare~\citep{claimverification-health}, finance~\citep{zhao2024findver} and scientific discovery~\citep{alvarez-etal-2024-zero}. In all settings, information underpinning the claims is crucial in ascertaining their veracity, but when this information is incomplete or conflicting,
\emph{uncertain answers} may be more appropriate than true or false classifications.
For example
FEVER \citep{Thorne18Fever} uses Supported, Refuted, and NotEnoughInfo labels, while AVeriTeC \citep{10.5555/3666122.3668964} distinguishes between Supported, Refuted, Not Enough Evidence, and Conflicting Evidence/Cherry-picking.
No matter the veracity labels, \emph{faithful explanations} of the considerations determining the labels are crucial in claim verification~\citep{10.1609/aaai.v39i14.33637}.

\begin{figure}
    \centering
    \includegraphics[width=0.89\linewidth]{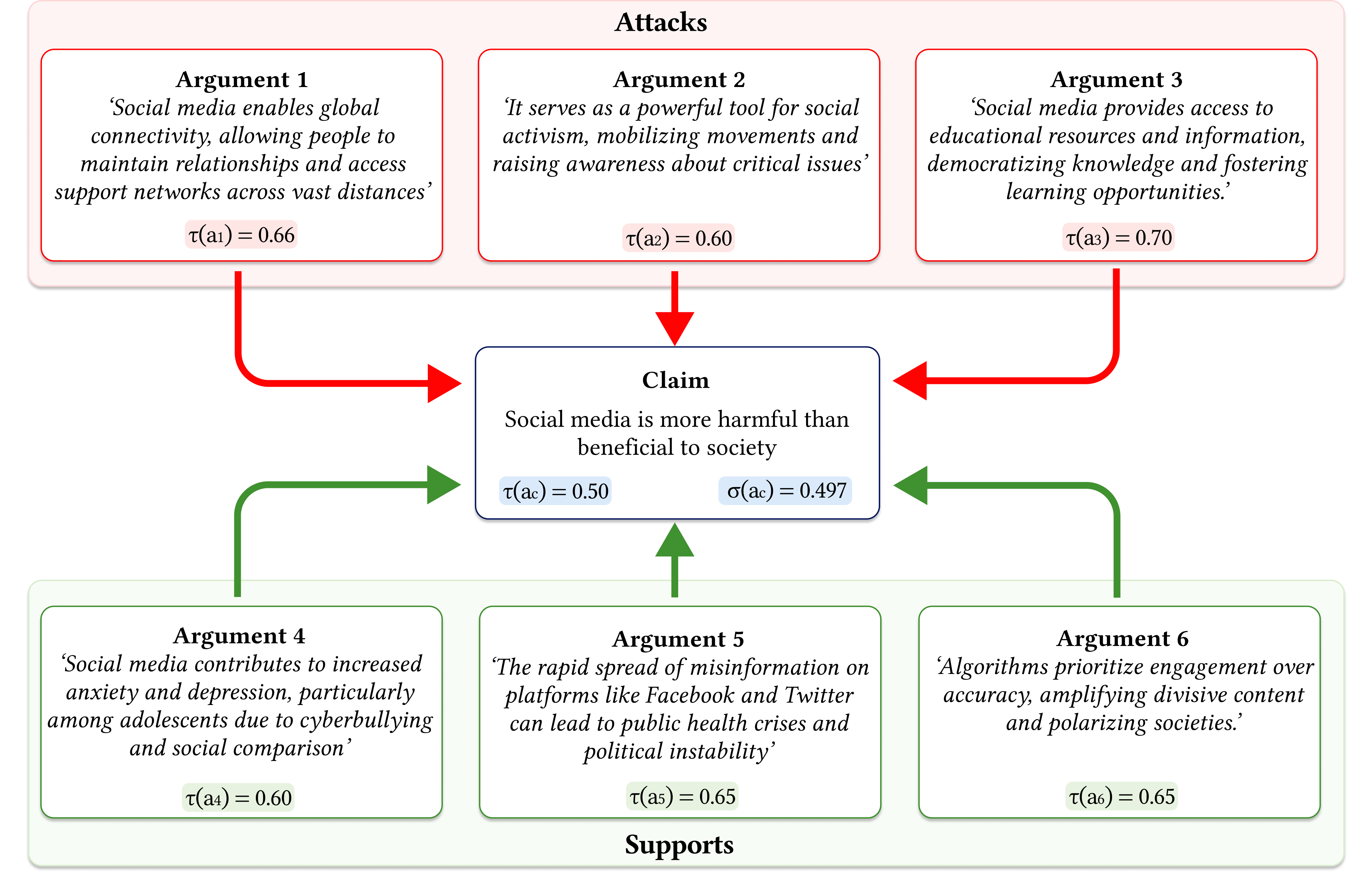}
      \caption{An example argumentative structure for a claim 
      adapted from DEBATunE~\citep{li-etal-2024-llms-speak}. 
      Arguments 1-3 attack the claim and 4-6 support it. For claim/argument $x$, $\tau(x)$ is the base score 
      and $\sigma(x)$ the argumentative strength computed by the DF-QuAD 
      semantics~\citep{Rago2016DiscontinuityFreeDS}. ITA classifies this claim as \emph{Uncertain} as its strength is close to the midpoint of [0,1]. 
      }
    \label{fig:example}
\end{figure}

In this paper 
we consider claim verification as a three-valued prediction problem. Given a natural language claim, we aim to automatically predict a verdict $\hat{y}\! \in \!\{\emph{True}, \emph{False}, \emph{Uncertain}\}$. The \emph{Uncertain} label is not treated as 
failure to classify, but as a substantive verdict for cases in which the available information does not warrant 
accepting or rejecting the claim, as 
in Figure~\ref{fig:example}.
To support this form of ternary claim verification while also guaranteeing faithful explanations for veracity verdicts, we introduce 
\emph{inference-time argumentation (ITA)},
a novel neurosymbolic framework for training LLMs to argue about the veracity of claims.
Rather than treating veracity outcomes as labels to be predicted directly, ITA derives them from the formal evaluation of generated supporting and attacking arguments
, making prediction an argumentative verdict rather than a bare classification.
In ITA,
similarly to 
ArgLLMs~\citep{10.1609/aaai.v39i14.33637},
verdicts are obtained by applying thresholds to a continuous \emph{argumentative strength} computed with a (deterministic) argumentation semantics~\citep{Rago2016DiscontinuityFreeDS}. The final verdicts are thus derived based on 
the argumentative structure of supporting and attacking arguments 
and the associated \emph{base scores} for all arguments and 
the claim, representing their intrinsic strength. 
As in ArgLLMs,
the argumentative structure serves as a faithful explanation for the verdict.
However,
rather than 
using off-the-shelf LLMs for generating and scoring arguments as ArgLLMs do~\citep{10.1609/aaai.v39i14.33637,Kevin25},
ITA is based on training them 
so that LLMs learn how to generate and/or score arguments optimally 
for the downstream prediction.
Thus, in ITA, 
the argumentation 
semantics also informs
training. Specifically, it supervises the arguments' base score generation 
and provides
feedback for argument generation with
reinforcement learning \CRchange{(RL)}. In other words, \CRchange{ITA is neurosymbolic in combining neural argument generation and scoring with symbolic reasoning over QBAFs. The formal argumentation semantics determines the verdict and provides the learning signals for the neural components.} 

ITA  differs from (non-argumentative) inference-time reasoning approaches based on 
reinforcement learning from verifiable rewards 
\citep{Guo_2025} in its focus on a task where outcomes may be uncertain, rather than exact and automatically checkable with certainty (e.g. as in mathematics or programming), and in its guarantees of faithful explainability, rather than with reasoning traces understood post-hoc as explanations,  
with several documented limitations~\citep{chen2025reasoningmodelsdontsay,cornish-rogers-2025-examining,yu2026outcomerewardsguaranteeverifiable}.

In summary, our contributions are as follows. First, we introduce 
ITA, a trainable neurosymbolic framework for three-valued claim verification in which verdicts are computed from 
argumentative structures (see Figure~\ref{fig:architecture}). Second, we introduce two 
training strategies for 
ITA, respectively based on (i)  
learning base scores for generated arguments through objectives defined over the downstream argumentative verdict, rather than relying only on base scores provided by base LLMs as in \cite{10.1609/aaai.v39i14.33637}, and
(ii) optimising the argument generator using feedback induced by formal argumentation semantics, turning argumentation from an inference-time decision layer into a training signal. 
Finally, we evaluate ITA on two claim-verification datasets against non-argumentative direct-prediction baselines and argumentative baselines
, showing that we outperform all faithfully explainable baselines while 
performing competitively against others.
We obtain the datasets by adapting three human-curated datasets: AVeriTeC~\citep{10.5555/3666122.3668964}, StrategyClaim~\citep{10.1609/aaai.v39i14.33637} and  DEBATunE~\citep{li-etal-2024-llms-speak}.

\section{Related Work}

\textbf{Abstention.}
Several LLM-based claim verification methods exist \citep{claimverification-survey}. Here we focus on methods
aware of their knowledge boundaries, 
that honestly say ``I do not know'' for non-answerable questions, 
known as abstention~\citep{li2024knowledge, wen2025know}.
Existing approaches teach LLMs to perform abstention through fine-tuning strategies~\citep{amayuelas2024knowledge, kapoorlarge2024}.
\citet{cohen2024don} further improve this line of research by introducing a special \texttt{[IDK]} (``\textit{I don't know}'') token into the model's vocabulary.
While these approaches all focus on training models to refuse to answer certain queries, they do not provide faithful explanations for why these refusals are the appropriate response. Nor do they seek to explicitly surface their reasoning when conflicting evidence, rather than just lack of knowledge, is the cause for their uncertainty.

\textbf{Learning and Argumentation.} Argumentation has been advocated in combination with learning in several settings (as recently surveyed by~\cite{maurizio23,Rago_24_HOFA}).
In the context of RL, 
\cite{ArgRL-Gao} use reasoning with argumentation frameworks to shape rewards, in a multi-agent simulated game setting, and 
\cite{francis-ICMLws2022} replace human
feedback in RL  with reasoning in preference-based argumentation, in a maze environment.
Neither of these approaches generates arguments or uses gradual argumentation semantics as we do.
In the context of neurosymbolic, end-to-end learning, \cite{nesy25Adam}
adopt gradual argumentation semantics as well as, like in our approach, 
base score model learning. However, in their approach arguments are given rather than generated by LLMs, and no RL  takes place.

\textbf{LLMs and Argumentation.}
Closest to our work are 
the proposals by \cite{10.1609/aaai.v39i14.33637} and \cite{ArgRAG-nico-nesy20-25}, both using argumentation in combination with LLMs, and highlighting its benefits in terms of faithfulness of explanations. However, both approaches focus on  generation of arguments and base scores from off-the-shelf LLMs rather than training them. Also,  both focus on binary rather than ternary claim verification.
%
Further, \cite{petrosAnsya25}
use a transformer-based encoder for feature extraction as an input to a symbolic learner for argumentation frameworks. However, the resulting neurosymbolic approach is not end-to-end. Moreover, the learning does not make use of argumentation semantics, focusing on a different form of (structured) argumentation.

\CRchange{\textbf{Explainable Claim Verification.}
PACAR~\citep{zhao-etal-2024-pacar} combines claim decomposition, dynamic planning and specialised reasoning actions, while LoCal~\citep{10.1145/3696410.3714748} uses agents for decomposition, reasoning and logical and causal evaluation. Both provide interpretable claim verification grounded in external evidence. However, their decision procedures still rely on LLM judgments over the generated reasoning. In ITA, the final verdict follows deterministically from the argumentative structure and base scores under the chosen formal semantics and thresholds. This guarantees procedural faithfulness by construction.} We leave the integration of external evidence as future work.
%



\section{Background}


\textbf{Quantitative Bipolar Argumentation Frameworks (QBAFs) and Argumentation Semantics.}
QBAFs~\citep{BARONI2019252} are graphical argumentative structures from the field of 
Argumentation~\citep{10.1609/aimag.v38i3.2704},
in which arguments are nodes and relations are edges 
representing support or attack relations, and  
arguments are assigned intrinsic strengths (referred to as \emph{base scores}). Formally, a QBAF is 
\(
Q = \langle A, R^{-}, R^{+}, \tau \rangle,
\)
where $A$ is a finite set of arguments, $R^{-} \subseteq A \times A$ is an attack relation, $R^{+} \subseteq A \times A$ is a support relation, and $\tau : A \rightarrow [0,1]$ assigns each argument a base 
score ($\langle A, R^{-}, R^{+}\rangle$ alone is called a BAF). 
Intuitively, $(a,b) \in R^{+}$ means that argument $a$ supports argument $b$, and $(a,b) \in R^{-}$ means that argument $a$ attacks argument $b$. 
Argumentation semantics assign 
\emph{(argumentative) strengths} to the arguments in a QBAF by aggregating their base 
scores with the effects of their supporters and attackers. Formally,
for a QBAF $Q$, 
an argumentation semantics is a function
\(
\sigma_Q: A \rightarrow [0,1]\).
Different 
semantics encode different assumptions about how support and attack should interact. \CRchange{In the main experiments,} we 
use DF-QuAD~\citep{Rago2016DiscontinuityFreeDS} 
for direct comparison with ArgLLMs which uses it already \citep{10.1609/aaai.v39i14.33637} (see Appendix \ref{aap:semantics} for the formal definition).


\textbf{Argumentative LLMs (ArgLLMs).} ArgLLMs \citep{10.1609/aaai.v39i14.33637} instantiate QBAFs and argumentation semantics  for binary (\emph{True/False}) claim verification.
They use
an LLM to construct a QBAF
\(
Q_c = \langle A_c, R_c^{-}, R_c^{+}, \tau_c \rangle
\)
for each given claim $c$. 
As arguments are represented in free text, we use the set $\mathcal{T}$ to refer to all possible sequences of input tokens and the set $\mathcal{A} \subseteq \mathcal{T}$ as the set of all possible arguments. Thus, we have that $A_c \subseteq \mathcal{A}$. 
In ArgLLMs, like in our approach,
the claim is represented as 
an argument $a_c \in A_c$, and generated arguments in $A_c \setminus \{a_c\}$ either support or attack $a_c$, as illustrated in Figure~\ref{fig:example}.
Then, an argumentation semantics $\sigma$ is applied to obtain the claim strength $\sigma_{Q
}(a_c)$\footnote{In the remainder, for readability, we will often omit the QBAF $Q$ and simply write $\sigma(a_c)$, as in Figure~\ref{fig:example}.}. A verdict is finally derived from this strength.
While ITA keeps the same argumentative structure (QBAFs) and semantics (DF-QuAD) as ArgLLMs, it differs by focusing on QBAFs of depth 1 only (ignoring QBAFs of depth 2) but with any number of attackers and supporters (rather than one attacker and one supporter). 
Also, while ArgLLMs are training-free, ITA learns how to generate arguments and how to assign base scores to them.


\textbf{Group Relative Policy Optimisation (GRPO).} GRPO \citep{shao2024deepseekmathpushinglimitsmathematical} is a critic-free method for RL
fine-tuning of LLMs, introduced as a resource efficient variant of Proximal Policy Optimisation. Given a prompt, the policy samples a group of completions, each of which receives a reward. Rather than learning a separate value model, GRPO estimates relative advantages by comparing rewards within the sampled group.
We adapt GRPO to argument generation, using rewards based on 
semantics.

\section{
Framework Overview}
\label{sec:method}

\begin{figure}
    \centering
    \includegraphics[width=1\linewidth]{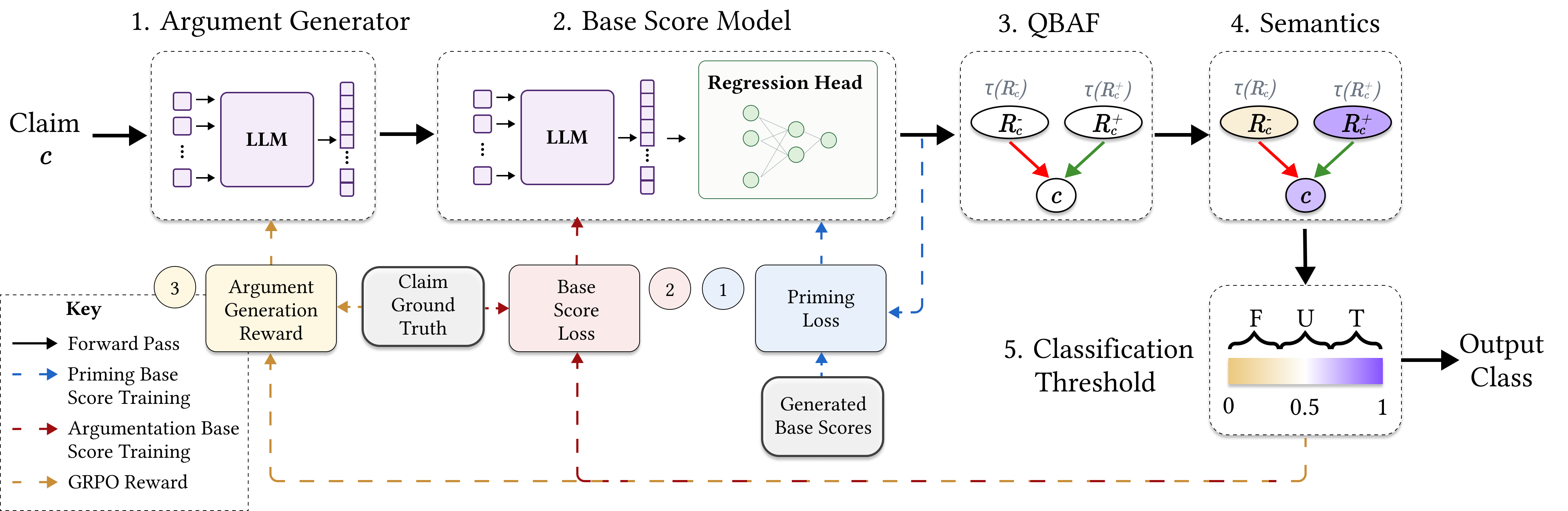}
    \caption{Overview of inference-time argumentation. Given a claim, the argument generator produces supporting and attacking arguments about the claim. 
    A base score module assigns intrinsic strengths to the generated arguments.
    These steps induce a QBAF centred on the 
    claim, which
    receives a fixed neutral base score of $0.5$. The regression head of the base score model can be primed using generated base scores during training (training phase \textcircled{1}). A
    n argumentation semantics 
    produces a final claim strength, which is thresholded into \emph{True}, \emph{False}, or \emph{Uncertain}. During training, the same semantic evaluation informs a weakly-supervised loss for training the base score model (training phase \textcircled{2}) and argumentation-based reward for the argument generator (training phase \textcircled{3}).} 
    \label{fig:architecture}
\end{figure}


The ITA framework consists of two trainable components (see Figure~\ref{fig:architecture}). The first is an \emph{argument 
generator}, parameterised by $\theta$, producing arguments supporting and attacking 
$c$:
\[
D_\theta(c) = (S_c, T_c),
\]
where $S_c$ and $T_c$ are  the set of generated supporting and attacking arguments
, respectively. These arguments induce a 
BAF \CRchange{$\langle A_c, R_c^{-}, R_c^{+}\rangle$} centred on the claim, 
$c$, 
represented as  
an argument $a_c$
, with:
\[
A_c = \{a_c\} \uplus S_c \uplus T_c,
\qquad
R_c^{+} = \{(a,a_c) : a \in S_c\},
\qquad
R_c^{-} = \{(a,a_c) : a \in T_c\}.
\]
The resulting graph represents the reasons offered for and against the veracity of the claim.

The second component is a \emph{base score model (BSM)}, parameterised by $\basescoreparam$, assigning 
base scores to the generated arguments.
This BSM is a learnable function
$\basescorefunc: \mathcal{A} \times \mathcal{A} \times \{0, 1\} \rightarrow [0, 1]$
that takes as input $a_c$, an argument $a$ and their relationship as given by the indicator function $\mathbf{1}_{a \in S_c}$ (returning $0$ for attacks and $1$ for supports), and outputs a 
value in $[0, 1]$.
The 
claim argument itself is not scored but instead its base score is fixed to a neutral prior of $0.5$. 
%
%
This gives the 
QBAF
\[
Q_c^{\theta,\phi}
=
\langle A_c,R_c^-,R_c^+,\tau_{\phi,c}\rangle,
\qquad
\text{with} \; \tau_{\phi,c}(a)=
\begin{cases}
0.5 & a=a_c,\\
\basescorefunc(a_c,a,\mathbf{1}_{a\in S_c}) & \text{otherwise}.
\end{cases}
\]
%
When the parameters are clear from context, we will write $Q_c$ for $Q_c^{\theta,\phi}$ and similarly suppress for readability the dependence on the parameters for both trainable components. An argumentation semantics $\sigma$ then aggregates the effects of supporting and attacking arguments, given their base scores, to compute the final strength of the 
claim argument:
\(
\hat v_c = \sigma(a_c) \in [0,1].
\)
This value represents the 
argumentative
strength of the claim after evaluating the generated debate.
It is mapped to a three-valued verdict $\hat{y}_c$ using \CRchange{thresholds $\alpha$ and $\beta$} by means of the following \emph{thresholding} function $h$:
\[
\hat{y}_c = h(\hat v_c) =
\begin{cases}
\emph{False} & \text{if } \hat v_c \leq \CRchange{\alpha}, \\
\emph{True} & \text{if } \hat v_c \geq \CRchange{\beta}, \\
\emph{Uncertain} & \text{otherwise.}
\end{cases}
\]

Thus, \emph{Uncertain} corresponds to an intermediate 
argumentative strength, arising when the debate represented by the QBAF does not make the claim sufficiently strong to accept or 
weak to reject.


\section{ 
Weakly Supervised Learning for Base Score Attribution}
\label{sec:bsm}

For our trained 
BSMs, we instantiate $\basescorefunc$ as follows:
\[
\basescorefunc(a_c, a, \mathbf{1}_{a \in S_c}) = \basescorehead(\basescorellm(g(a_c, a, \mathbf{1}_{a \in S_c})))
\]
We use three components.
First, $g: \mathcal{A} \times \mathcal{A} \times \{0, 1\} \rightarrow \mathcal{T}$ constructs a task-specific prompt\footnote{Provided in Appendix~\ref{app:base_score_prompt}.} from the claim, the argument and their relationship. Second, $\basescorellm: \mathcal{T} \rightarrow \mathbb{R}^{d}$ maps the prompt to a $d$-dimensional latent space embedding using an LLM. Finally, $\basescorehead: \mathbb{R}^{d} \rightarrow [0, 1] $ is a feed-forward regression head that converts the embedding to a scalar score. 
Intuitively, the prompt generator encodes the scoring task, the LLM produces a contextual embedding of the argument relative to the claim, and the regression head maps this embedding to a base score.

Then, the BSM is trained in two phases. In the first phase (\textcircled{1} in Figure~\ref{fig:architecture}) the regression head is primed with no regard for the prediction task. Thus, the parameters of $\basescorellm$ are frozen and only the parameters of $\basescorehead$ are allowed to update. In this phase, claim, argument and base score triples $(a_c, a, t)$ 
are used in a supervised manner to update the regression head. In all our experiments, the target base scores $t$ are generated by directly prompting the base LLM, optimising the regression head to use argument quality information already reflected in the model embedding.
For each claim-argument pair, we compute $\hat{t} = \basescorefunc(a_c, a, \mathbf{1}_{a \in S_c})$. Using a standard gradient-based optimiser, we can update the parameters $\basescoreparamhead$ using a mean square error loss function$$\mathcal{L}_{primed}(\hat t, t) = \mathcal{L}_{MSE}(\hat t, t).$$ This step is shown in Figure~\ref{fig:architecture} as the priming loss. 

In the second phase of training (\textcircled{2} in Figure~\ref{fig:architecture}), all trainable model parameters can freely update. In this phase, the BSM is trained using weak supervision for the claim verification task. Each step of training requires a provided claim, sets of supporting and attacking arguments, and the ground-truth labelling of the claim. For each argument a base score is computed with $\basescorefunc$ followed by application of the argumentation semantics to give a 
strength for the claim. As 
DF-QuAD 
is differentiable, we can back-propagate through 
it to update the parameters of the BSM. The loss function is $$\mathcal{L}_{sem}(\hat v_c, v_c) = \mathcal{L}_{MSE}(\hat v_c, v_c)$$ where $\hat v_c = \sigma(a_c)$ and $v_c \in \{0, 0.5, 1\}$ is a mapping of the ground truth label of the claim: 
\emph{True} to 1, \emph{False} to 0, and \emph{Uncertain} to 0.5. This step is shown in Figure~\ref{fig:architecture} as the base score loss. Note that this is a neurosymbolic semantic loss function using argumentation constraints rather than logical constraints to weakly supervise the BSM. 

Additional loss terms can be used in combination with this 
semantic loss to optimise for desired behaviours of the BSM. We propose two such behaviours: (i) consistency in scoring semantically similar arguments and, (ii) making the BSM-induced ranking of arguments conform to some provided ordering.  
For (i)
, 
the consistency loss is
$$\mathcal{L}_{con}(a_c, \mathcal{C}) = \frac{1}{n} \sum_{i =1}^{n}(t_i - \bar{t})^2$$ where $\mathcal{C} = (a_1, a_2, \ldots, a_n) $ is a tuple of arguments generated for 
$a_c$ assumed to be semantically equivalent (i.e. paraphrases), $t_i = \basescorefunc(a_c, a_i, \mathbf{1}_{a_i \in S_c})$ and $\bar{t} = \frac{1}{n} \sum t_i$. In effect, the goal of 
this loss is to minimise the variance of computed base scores for arguments that are semantically similar. However, $\mathcal{C}$ must be pre-generated for all input claims, which may result in a large number of samples 
not feasible for training. Thus, 
we generate $\mathcal{C}$ for only a subset of claims and then 
randomly sample claim-tuple pairs 
therefrom and compute the associated loss.
For (ii), 
the ranking loss is
$$\mathcal{L}_{rank}(a_c, \mathcal{P}) = \frac{2}{n \times (n-1)} \sum_{i =1}^{n} \sum_{j= i + 1}^{n}  \max(0, m - (t_i - t_j))$$
where $\mathcal{P} = (a_1, a_2, \ldots, a_n)$ is a tuple of arguments generated for 
$a_c$ in ranked order where the base score of $a_i$ is expected to be greater than $a_j$ for $i < j$ and, as before, $t_i = \basescorefunc(a_c, a_{i}, \mathbf{1}_{a_i \in S_c})$. In effect, this is a pairwise margin ranking loss that encourages the base scores to conform to a predefined ordering and to be separated by at least a margin of $m$, which is a hyperparameter. Again, as $\mathcal{P}$ must be pre-generated, we randomly sample claim-tuple pairs to compute this loss.

The full loss in phase 2 can be summarised as: 
$$\mathcal{L} = \mathcal{L}_{sem} + \lambda_{rank} \mathcal{L}_{rank} + \lambda_{con} \mathcal{L}_{con}\footnote{The three constituent losses are estimated from separate data resources. The fully specified loss is given in Appendix~\ref{app:bsm-training}.}$$
where $\lambda_{rank}$ and $\lambda_{con}$ are hyperparameters that weigh their respective loss functions.






\section{Reinforcement Learning for Argument Generation}
\label{sec:rl}

The argument generat
or
is trained using a GRPO procedure with the aim of encouraging the generation of arguments helpful for the downstream claim verification task. The training process takes 
a training dataset $\mathcal{D} = \{(c_i, y_i)\}_{i = 1}^N$ composed of claims $c_i$ and corresponding 
ground-truth labels $y_i$, 
an \emph{initial argument generator} $D_{\theta_{\text{init}}}$, a fixed 
BSM $f_\phi$ and an argumentation semantics $\sigma$. During training, each claim in the dataset is passed to the \emph{current argument generator} $D_{\theta_{\text{curr}}}$ to generate the associated sets of supporting and attacking arguments. These arguments are then scored by the BSM, organised with their respective claim into a QBAF and evaluated by 
$\sigma$, yielding final strengths 
for the claims
, as described in Section \ref{sec:method}. For each claim, we sample $G$ outputs from 
$D_{\theta_{\text{curr}}}$
to obtain the group of samples required by GRPO. For each obtained claim strength $\hat{v}_c$, we compute a scalar reward $r_\rev{\alpha, \beta}(y, \hat{v}_c)$, with $(c,y)\in \mathcal{D}$, using our custom \emph{reward function}, which is defined, for \rev{$0 < \alpha < \beta < 1$}, as:
\begin{align*}
    r_\rev{\alpha, \beta}(\emph{True}, \hat{v}_c) &= \begin{cases}
        100 & \text{if $\hat{v}_c \geq \rev{\beta}$} \\
        200 \times \frac{\hat{v}_c}{\rev{\beta}} - 100 & \text{otherwise}
    \end{cases} \\
    r_\rev{\alpha, \beta}(\emph{False}, \hat{v}_c) &= \begin{cases}
        100 & \text{if $\hat{v}_c \leq \rev{\alpha}$} \\
        200 \times \frac{1 - \hat{v}_c}{1 - \rev{\alpha}} - 100 & \text{otherwise}
    \end{cases} \\
    r_\rev{\alpha, \beta}(\emph{Uncertain}, \hat{v}_c) &= \begin{cases}
        100 & \text{if $\rev{\alpha \leq \hat{v}_c \leq \beta}$} \\
        \rev{100 - 200 \times \frac{\alpha - \hat{v}_c}{\alpha}} & \rev{\text{if $\hat{v}_c < \alpha$}} \\
        100 - 200 \times \frac{\rev{\hat{v}_c - \beta}}{\rev{1 - \beta}} & \text{otherwise}
    \end{cases}
\end{align*}
\rev{Note that $r_{\alpha, \beta}$ as defined is consistent with our thresholding function.}
This \rev{means that it provides} a maximum reward for any claim strength that would result in the correct classification while assigning lower rewards to scores that would result in prediction errors, decreasing linearly with the increasing distance to the corresponding threshold. \rev{Model completions that cannot be parsed into supporting and attacking arguments receive a fixed negative reward of $-1000$.} The reward is used with the standard GRPO loss function \citep{shao2024deepseekmathpushinglimitsmathematical} during training.

\section{Data}

\CRchange{Training uses claim-level labels together with generated supporting and attacking arguments. The semantic loss is supervised by ground-truth claim labels, while generated base scores are used for priming, and synthetic paraphrases and rankings support the optional auxiliary losses.} We use separate data resources for training, validation, and final evaluation, with full construction details provided in Appendix~\ref{app:Data}. Training and validation include LLM-generated arguments, while final evaluation is conducted on claims drawn from existing human-curated datasets (where necessary minimally modified to become appropriate for stand-alone claim verification). We keep the two evaluation datasets strictly separate from data used in any training resource.

For training, we construct \emph{TFU Training Claims}, a set of 1{,}000 claims evenly split between labels \emph{True}, \emph{False}, or \emph{Uncertain}. The \emph{True} and \emph{False} claims are drawn from TruthfulClaim \citep{10.1609/aaai.v39i14.33637}, while the \emph{Uncertain} claims are drawn from the Rational Probabilistic Beliefs dataset \citep{freedman2025exploring}. For training the BSMs, we generate two sets of supporting and attacking arguments for a subset of 
the claims in \emph{TFU Training Claims}, using GPT-5 \citep{singh2025openai} and Qwen3-8B \citep{yang2025qwen3} respectively
.  We also generate ranking and consistency datasets for the respective losses described in Section~\ref{sec:bsm}. The consistency dataset is generated from a subset of the Qwen3-8B arguments based on \emph{TFU Training Claims}, by generating three semantically-equivalent paraphrases of each argument with GPT-5. The ranking dataset is also based on a subset of \emph{TFU Training Claims}, and uses GPT-5 to generate four candidate arguments in descending order of quality. 

The validation data is composed of 150 synthetically generated claims (using GPT-5 and Claude Sonnet 4.5), evenly split across the three classes, together with supporting and attacking arguments generated by Qwen3-8B.

We evaluate on two ternary claim-verification datasets. The first, \emph{TFU Evaluation Claims}, contains 750 claims balanced across \emph{True}, \emph{False}, and \emph{Uncertain}. The \emph{True} and \emph{False} claims are drawn from StrategyClaim \citep{10.1609/aaai.v39i14.33637}, while the \emph{Uncertain} claims are drawn from DEBATunE \citep{li-etal-2024-llms-speak}. Where necessary, DEBATunE questions are converted into declarative factual claims using GPT-5, as well as filtered to exclude statements of opinion using simple string-matching (e.g. excluding any sample containing `should'), and human  verification. The second evaluation set contains 1{,}746 real-world claims from AVeriTeC \citep{10.5555/3666122.3668964}. Where the claims have an ambiguous temporal component, we use GPT-5 to add the date provided in the metadata to the claim, forming a single sentence. Since the resulting AVeriTeC split is imbalanced, with 334 \emph{True}, 1{,}190 \emph{False}, and 222 \emph{Uncertain} claims, we report balanced accuracy.

\section{Experiments}
\label{sec:experiments}


We evaluate whether training the components of ITA improves ternary claim verification while preserving the constraint that predictions are computed from explicit argumentative structures. \CRchange{The main experiments} use Qwen3-8B \citep{yang2025qwen3} as the underlying LLM. We use Qwen3-8B because it is a strong open-weight model that fits our compute budget and is supported by our fine-tuning stack, allowing all direct and argumentative variants to be compared under the same base model. Training is performed using low-rank adaptation (LoRA) parameter-efficient fine-tuning where applicable \citep{hu2022lora}. All argumentative variants (baselines and ours) \CRchange{in the main comparison} use fixed thresholds at \CRchange{$\alpha = 1/3$ and $\beta = 2/3$} and DF-QuAD semantics. Furthermore, the argumentative baselines and trained 
BSM 
experiments \CRchange{in the main comparison} all use the same argumentative structures generated by an untuned Qwen3-8B model, to make sure the results are only measuring variation in base score attribution (but naturally in the experiments with trained argument generators 
we do not impose this restriction).

We report accuracy on  \emph{TFU Evaluation Claims} and balanced accuracy on the AVeriTeC evaluation set. Further implementation details, including hyperparameters, are given in Appendix~\ref{app:hyperparams}.

\subsection{Experimental Variations}

We experiment with five classes of systems, as in Table~\ref{tab:variant-accuracy}: \emph{Non-Argumentative Baselines}, \emph{Argumentative Baselines}, \emph{Argument Generator Model}, \emph{Base Score Models} and \emph{Combination of Models}.

\textbf{Non-Argumentative Baselines.}  The first variety of baselines classifies claims without constructing 
QBAFs. These include the untuned Qwen3-8B model, a supervised fine-tuned direct predictor (SFT), and a direct GRPO model trained with reward on the final label. These baselines test whether ITA provides benefits beyond standard prompting, supervised fine-tuning, or reinforcement learning on labels. Full details of the baselines can be found in Appendix~\ref{app:baselines}.

\textbf{Argumentative Baselines.} The second variety of baselines constructs QBAFs but does not train ITA components. In these variants, arguments are generated by Qwen3-8B and base scores are either fixed to $0.5$ or assigned by prompting Qwen3-8B. These baselines isolate the effect of formal argumentative inference without training, as in ArgLLMs \citep{10.1609/aaai.v39i14.33637}.

\textbf{Base Score Models.} These variants keep the evaluation-time arguments fixed to the same Qwen3-8B generations as in \emph{Argumentative Baselines}, but replace constant or prompted base scores with a learned base score model. This isolates the effect of training argument scores through the downstream argumentative verdict. We consider 
BSMs 
trained on Qwen3-8B-generated arguments (BSM-Q) and GPT-5-generated arguments (BSM-G), using either the semantic loss alone or the semantic loss with ranking and consistency auxiliaries, as described in Section~\ref{sec:bsm}.

\textbf{Argument Generator Model.} Here, the argument generator is optimised with GRPO using argumentation-based feedback (see Section~\ref{sec:rl}
), while base scores are assigned by the prompted untuned Qwen3-8B scorer. This isolates whether reinforcement learning improves the generated argumentative structure independently of the learned base score module.

\textbf{Combination of Models.} These variants pair the GRPO-trained argument generator with a learned BSM. We select one BSM-Q and one BSM-G variant based on validation performance, testing whether trained generation and trained scoring provide complementary gains. \CRchange{The BSM is trained first and remains fixed during GRPO training of the argument generator.}

\subsection{Main Performance Results}

\begin{table}[tb!]
\centering
\caption{Main performance results (accuracy on TFU and balanced accuracy on AVeriTeC). For the argumentative baselines and our models, A+B indicates which method A was used for argument generation (args) and which method B for generation of base scores. `BSM-Q' stands for `base score model trained on arguments generated by Qwen3-8B', and `BSM-G' stands for `base score model trained on arguments generated by GPT-5'. `sem', `rank' and `con' refer to different losses (see Section~\ref{sec:bsm}). The best results overall are bolded while the best argumentative method results are underlined.
See Table \ref{tab:variant-accuracy-cis} in the SM for further results and confidence intervals.}
\label{tab:variant-accuracy}
\begin{tabular}{lcc}
\toprule
Variant & TFU & AVeriTeC \\
\midrule
\multicolumn{3}{l}{\textit{Non-Argumentative Baselines}} \\
\midrule
Qwen direct baseline
  & 0.635 & 0.454 \\
Qwen direct SFT
  & 0.659 & 0.446 \\
Qwen direct GRPO
  & 0.592 & \textbf{0.485} \\
\midrule
\multicolumn{3}{l}{\textit{Argumentative Baselines}} \\
\midrule
Qwen args + constant 0.5 base scores
  & 0.555 & 0.405 \\
Qwen args + Qwen base scores
  & 0.585 & 0.406 \\
\midrule
\multicolumn{3}{l}{\textit{Argument Generator Model (Ours)}} \\
\midrule
Qwen GRPO args + Qwen base scores
  & 0.591 & 0.426 \\
\midrule
\multicolumn{3}{l}{\textit{Base Score Models (Ours)}} \\
\midrule
Qwen args + BSM-Q (primed \& sem)
  & 0.588 & 0.415 \\
Qwen args + BSM-Q (primed \& sem \& rank \& con)
  & 0.590 & \underline{0.471} \\
Qwen args + BSM-G (primed \& sem)
  & \textbf{\underline{0.677}} & 0.460 \\
Qwen args + BSM-G (primed \& sem \& rank \& con)
  & 0.584 & 0.466 \\
\midrule
\multicolumn{3}{l}{\textit{Combination of Models (Ours)}} \\
\midrule
Qwen GRPO args + BSM-Q (primed \& sem \& rank \& con)
  & \CRchange{0.581} & 0.468 \\
Qwen GRPO args + BSM-G (primed \& sem)
  & 0.668 & 0.458 \\
\bottomrule
\end{tabular}
\end{table}

Table~\ref{tab:variant-accuracy} reports the main results. The non-argumentative (direct-prediction) baselines are strong, with supervised fine-tuning performing best on TFU and direct GRPO performing best on AVeriTeC. The overall strong performance is expected as these models are optimised to predict labels directly, without requiring the prediction to be mediated by generated arguments and formal semantics.

The training-free argumentative baselines perform below the direct baselines. Constant base scores test whether the generated graph structure alone is sufficient, while prompted Qwen scores test whether an untuned LLM can assign useful intrinsic strengths.

The learned 
BSMs provide the clearest gains among the argumentative 
variants. On TFU, the best learned scorer improves over both argumentative baselines and the strongest direct-prediction baseline. On AVeriTeC, learned base scores substantially improve over constant and prompted argumentative scoring, and approach the strongest direct baseline. In particular, the best-performing BSM on the validation data, `BSM-G (primed \& sem)', significantly outperforms argumentative baselines on both test datasets, as indicated by the confidence intervals in Table \ref{tab:variant-accuracy-cis}. These results support the central claim that formal argumentation semantics can be used as a useful training signal for argument scoring, rather than only as an inference-time aggregation rule.

\CRchange{The priming-only results in Table~\ref{tab:variant-accuracy-cis} further isolate the contribution of semantic training. For BSM-G, adding the semantic loss improves TFU accuracy from $0.588$ to $0.677$ and AVeriTeC balanced accuracy from $0.410$ to $0.460$. The auxiliary ranking and consistency losses improve the local properties examined in Table~\ref{tab:bsm-diagnostics}, but their effects on downstream accuracy vary across datasets and training arguments.}

The GRPO-trained argument generator gives mixed but informative results. When paired with prompted Qwen3-8B scores, it improves slightly over the corresponding Qwen3-8B-generated argumentative baseline on both datasets, suggesting that argumentation-based feedback can shape generation. However, the current results indicate that generator training is less consistently beneficial than base score training. This suggests that argumentation-based reward can shape generation in a useful direction, but the effect is modest compared with base score learning.

Finally, combining GRPO training for argument generation with the best learned BSMs does not yield additive improvements. The combined systems remain competitive with the strongest argumentative models, but slightly underperform the corresponding BSMs evaluated on the original Qwen3-8B-generated arguments. This suggests that improving argument generation and improving argument scoring are not independent plug-in gains. \CRchange{One possible explanation is a mismatch between the arguments used to train the fixed scorer and those produced by the GRPO-trained generator.} Under the current sequential pipeline, base score learning is thus the more reliable component. Future work should investigate joint or alternating optimisation of generator and scorer.

\subsection{Additional BSM Experiments}

\begin{table}[tb]
\centering
\caption{Diagnostic results for base score estimation strategies. Inconsistency reports the variance in scores assigned to semantically equivalent arguments (
lower is better). Ranking reports Spearman rank agreement with the expected ordering of 
strengths 
(
higher is better). 
Best results 
are 
in bold.}
\label{tab:bsm-diagnostics}
\begin{tabular}{lcc}
\toprule
Variant & Inconsistency ($\downarrow$) & Ranking ($\uparrow$) \\
\midrule
\multicolumn{3}{l}{\textit{Argumentative Baseline}} \\
\midrule
Qwen args + Qwen base scores
  & 0.013 & 0.878  \\
\midrule
\multicolumn{3}{l}{\textit{Base Score Models (Ours)}} \\
\midrule
Qwen args + BSM-Q (primed \& sem)
  & 0.007 & 0.525  \\
Qwen args + BSM-Q (primed \& sem \& rank \& con)
  & \textbf{0.005} & \textbf{0.944}  \\
Qwen args + BSM-G (primed \& sem)
  & 0.008 & 0.709  \\
Qwen args + BSM-G (primed \& sem \& rank \& con)
  & 0.006 & 0.921 \\
\bottomrule
\end{tabular}
\end{table}

We conducted 
further experiments to measure whether 
BSM training can impart desirable argumentative properties beyond downstream accuracy: consistency in scores assigned to equivalent arguments and correct ranking of arguments gradually decreasing in quality. 
The results are summarised in Table~\ref{tab:bsm-diagnostics} and 
discussed in Appendix~\ref{app:additional-experiments}. \CRchange{Since no unique gold intrinsic strength is assumed, these diagnostics provide a functional assessment of the learned base scores.}

\CRchange{Further experiments examine threshold selection (Appendix~\ref{app:thresholds}), alternative argumentation semantics (Appendix~\ref{app:qe}) and a second LLM backbone (Appendix~\ref{app:gemma}).}

\section{Conclusion}

We introduced \emph{inference-time argumentation} (ITA), a trainable neurosymbolic framework for three-valued claim verification 
extending the role of argumentation from an inference-time explanation layer into a learning interface for LLM-based claim verification.
Our experiments show that training the argumentative components, especially the base score model, can benefit downstream performance. Learned base score models improve over argumentative baselines, and auxiliary consistency and ranking objectives improve desirable local properties of the scores. 
The resulting explanations are faithful by construction, as the final verdict is derived from the generated debate, the learned base scores, and the chosen argumentation semantics, rather than justified by a post-hoc reasoning trace.

\paragraph{Future Work.} \CRchange{ITA currently reasons over generated arguments without requiring them to be grounded in retrieved evidence. For practical claim verification, future work should integrate evidence retrieval so that generated support and attacks are tied to external sources rather than only to model knowledge, potentially drawing on PACAR~\citep{zhao-etal-2024-pacar} and LoCal~\citep{10.1145/3696410.3714748}.} Another important direction is to move beyond the single layer QBAFs used here, allowing arguments to support or attack one another and thereby represent richer debates. Future work could also conduct human-centred evaluations of the contestability offered by ITA, in the spirit of \cite{10.1609/aaai.v39i14.33637}. \CRchange{Such evaluations could assess preferences for arguments generated before and after GRPO, agreement between BSM scores and human judgments, and whether ITA explanations aid human decision-making.} \CRchange{Finally, beyond the targeted threshold studies reported here, we leave a broader exploration of ITA's hyperparameters to future work.}  


\begin{ack}
Freedman was funded by UKRI through the CDT in Safe and Trusted Artificial Intelligence (Grant No. EP/S023356/1). Dejl, Chen and Toni were partially supported by ERC under the EU's Horizon 2020 research and innovation programme (grant agreement No. 101020934, ADIX). Toni was also funded by EPSRC (grant UKRI3928, NeSyDebates). Gould was supported by UK Research and Innovation [UKRI Centre for Doctoral Training in AI for Healthcare grant number EP/S023283/1]. Finally, Mansi was supported by the Imperial College Departmental Scholarship. 
All authors are grateful to the other members of the CLArg group at Imperial for useful discussions and feedback. There are no competing interests to declare.
\end{ack}

\bibliographystyle{plainnat}
\bibliography{references}


\appendix

\clearpage

\section{Prompts}
\label{app:prompts}

The prompts shown below are used for all variations of the respective components.

\subsection{Argument Generation}
\label{app:arg_gen_prompt}

\begin{quote}
\ttfamily\small
Please provide a set of short arguments supporting and attacking the following claim.\\
Construct the arguments so they refer to the truthfulness of the claim.\\
The arguments should be short and concise, ideally one sentence per argument.\\
The arguments should be based on factual information and logical reasoning.\\
Only provide supporting or attacking arguments if you think there is a valid and convincing support or attack for this claim, meaning there is a non-zero probability that the claim is true or false; otherwise return: N/A.\\
Provide as many unique supporting and attacking arguments as you think are correct and applicable to the claim.\\[0.75em]

Respond in the following format:\\[0.5em]

claim:\\
\hspace*{2em}``\{claim\}''.\\[0.5em]

Output:\\
\{\\
\hspace*{2em}'support': [\\
\hspace*{4em}``<SUPPORT ARGUMENT 1>'',\\
\hspace*{4em}``<SUPPORT ARGUMENT 2>'',\\
\hspace*{4em}\ldots\\
\hspace*{2em}],\\
\hspace*{2em}'attack': [\\
\hspace*{4em}``<ATTACK ARGUMENT 1>'',\\
\hspace*{4em}``<ATTACK ARGUMENT 2>'',\\
\hspace*{4em}\ldots\\
\hspace*{2em}]\\
\}\\[0.75em]

Claim: \{claim\}
\end{quote}

\subsection{Base Score Model}
\label{app:base_score_prompt}

\begin{quote}
\ttfamily\small
You are an analyst evaluating the validity and relevance of arguments.\\[0.5em]

For the argument:\\[0.5em]

Argument: ``\{statement\}''\\[0.5em]

Please give your confidence that the argument presents a compelling case \{in favour of / against\} the statement:\\[0.5em]

Statement: ``\{claim\}''\\[0.5em]

Your assessment should be based on how well the argument \{supports / refutes\} the considered statement, as well as the correctness, accuracy, and truthfulness of the given argument.\\
Your response should be between 0\% and 100\%, with 0\% indicating that the considered argument is definitely invalid, 100\% indicating that the considered argument is definitely valid, and values in between indicating various levels of uncertainty.\\
Your estimates should be well-calibrated, so feel free to err on the side of caution and output moderate probabilities if you are not completely sure in your assessment.\\
Please respond in the following form:\\[0.5em]

[The predicted likelihood that the considered argument is valid as a number between 0 and 100]\%\\[0.5em]

Reply only with the predicted likelihood without any additional text or explanation.
\end{quote}

\subsection{Baselines}
\label{app:prompt_baseline}

\begin{quote}
\ttfamily\small
You are a careful fact-checking assistant.\\
Classify the claim below using exactly one of the following three choices:
True, False, or Uncertain.\\
Return only your choice as a single word and nothing else.\\[0.75em]
Claim: \{claim\}
\end{quote}

\section{Datasets}
\label{app:Data}

We use separate data resources for claim labels, argument generation, base score training, and final evaluation. Training and validation partially consist of synthetic arguments produced by LLMs, while final evaluation is conducted on claims drawn from existing human-curated datasets. No evaluation claims are used to construct the training resources.

\subsection{Training}
\label{app:main_training_data}
\paragraph{TFU Training Claims.}
We construct a training set of 1{,}000 claims with labels in $\{\emph{True}, \emph{False}, \emph{Uncertain}\}$. The set contains 333 \emph{True} claims, 333 \emph{False} claims, and 334 \emph{Uncertain} claims. The \emph{True} and \emph{False} claims are drawn from TruthfulClaim \citep{10.1609/aaai.v39i14.33637}, while the \emph{Uncertain} claims are drawn from the Rational Probabilistic Beliefs dataset \citep{freedman2025exploring}.

\paragraph{TFU Training Arguments.}
For \rev{1000} claims from TFU Training Claims, we generate supporting and attacking arguments using \rev{Qwen3-8B}. \rev{Separately, we also generate arguments} for \rev{499} of these claims with \rev{GPT-5}. These claim-argument sets are used to train the base score model components.

\paragraph{Consistency Training Arguments.} We construct 2{,}330 paraphrase groups covering 284 claims from \emph{TFU Training Arguments}. Each group contains three semantically equivalent rewordings, generated by GPT-5, of a single argument. These groups are used to train the base score model to assign consistent scores to semantically equivalent arguments.

\paragraph{Ranking Training Arguments.} We construct ranking data for 493 claims from \emph{TFU Training Claims}. Each instance contains a list of four arguments ranked from strongest to weakest with respect to a fixed claim and relation type, giving 164 supporting sets and 329 attacking sets. Arguments are generated with GPT-5 and are then filtered by twice providing them as input to Claude Sonnet 4.5 in a random order, and removing any that are not returned in the `correct' generated order on both occasions. These data are used to train the base score module to respect relative argument strength.

\subsection{Validation}

\paragraph{TFU Validation Claims.} We use 150 validation claims generated synthetically by Claude Sonnet 4.5 and GPT-5, evenly split across \emph{True}, \emph{False}, and \emph{Uncertain}.

\paragraph{TFU Validation Arguments.}
For each validation claim, we generate supporting and attacking arguments using Qwen3-8B, yielding 150 validation claim-argument sets.

\subsection{Evaluation}
\label{app:dataset_eval}

Final evaluation is conducted on claims drawn from existing human-curated datasets. Where necessary, we use LLM-assisted rewriting only to convert source items into declarative factual claims.

\paragraph{TFU Evaluation Claims.}
We construct a balanced evaluation set of 750 claims, with \rev{249, 251, and 250} \emph{True}, \emph{False}, and \emph{Uncertain} \rev{claims, respectively}. The \emph{True} and \emph{False} claims are drawn from StrategyClaim \citep{10.1609/aaai.v39i14.33637}, while the \emph{Uncertain} claims are drawn from DEBATunE \citep{li-etal-2024-llms-speak}. Where necessary, questions from DEBATunE are transformed into declarative factual claims using GPT-5 and filtered to exclude opinion statements. For example, ``The use of robotics in manufacturing leads to better product quality.'' is retained as a factual claim, whereas statements involving prescriptions such as ``We should ban lotteries.'' are excluded.

\paragraph{TFU Evaluation Arguments.}
For each TFU Evaluation Claim, we generate supporting and attacking arguments using Qwen3-8B, yielding 750 claim-argument sets. These are used to evaluate the base score model on arguments generated by the same model family used at test time.

\paragraph{AVeriTeC Claims.}
We also evaluate on 1{,}746 real-world claims drawn from AVeriTeC \citep{10.5555/3666122.3668964}. The resulting three-valued set is class imbalanced, with 1{,}190 \emph{False}, 334 \emph{True}, and 222 \emph{Uncertain} claims, so we report balanced accuracy. For time-sensitive claims, we use the metadata accompanying the dataset to make the temporal context explicit without adding external information. For example, ``All government schools in India are being privatised.'' is rewritten as ``As of August 22, 2020, all government schools in India are being privatised.''. To assign labels, we map \textsc{Supported} to \emph{True}, \textsc{Refuted} to \emph{False}, and \textsc{Conflicting Evidence/Cherry-picking} to \emph{Uncertain}. \textsc{Not Enough Evidence} claims are disregarded, as we are more interested in claims that are uncertain due to conflicting information rather than just a lack of evidence.

\paragraph{AVeriTeC Arguments.}
For each AVeriTeC claim, we generate supporting and attacking arguments using Qwen3-8B, yielding 1{,}746 claim-argument sets.

\paragraph{Consistency Evaluation Arguments.}
We generate 150 sets of paraphrases for evaluation, using the same methodology as \emph{Consistency Training Arguments} in Appendix~\ref{app:main_training_data}, but instead using \emph{TFU Evaluation Arguments}.

\paragraph{Ranking Evaluation Arguments.}
Likewise, we construct an additional \rev{149} sets of ranking arguments using the same methodology as \emph{Ranking Training Arguments}, but based on \emph{TFU Evaluation Claims}.

\begin{table}[t]
\centering
\small
\caption{Summary of data resources used for training, validation, and evaluation.}
\begin{tabular}{llll}
\toprule
Resource & Size & Provenance & Use \\
\midrule
TFU Training Claims & 1{,}000 & TruthfulClaim, RPB & Claim-label training \\
TFU Training Arguments & \CRchange{\rev{499/1000 sets}} & \CRchange{GPT-5, Qwen3-8B} & Argumentative training \\
Consistency Training & 2{,}330 groups & GPT-5 & base score consistency \\
Ranking Training & 493 groups & GPT-5, Claude Sonnet 4.5 & base score ranking \\
TFU Validation Claims & 150 & GPT-5, Claude Sonnet 4.5 & Validation \\
TFU Evaluation Claims & 750 & StrategyClaim, DEBATunE & Final evaluation \\
AVeriTeC Claims & 1{,}746 & AVeriTeC & Final evaluation \\
\bottomrule
\end{tabular}
\label{tab:data-summary}
\end{table}

\section{Direct Prediction Baselines}
\label{app:baselines}

We compare our argumentative models against direct prediction baselines that use the same underlying LLM but do not construct an argumentation framework. These baselines are intended to isolate the contribution of inference-time argumentation from the benefits of standard prompting, supervised fine-tuning, and reinforcement learning on the final label. In all direct baselines, the model is given a claim $c$ and is prompted to output exactly one label from $\{\emph{True}, \emph{False}, \emph{Uncertain}\}$.

\paragraph{Qwen Direct Baseline.}
The first baseline is an untuned direct predictor. We prompt Qwen3-8B to classify each claim directly as \emph{True}, \emph{False}, or \emph{Uncertain}. At evaluation time, generations are decoded greedily and the first valid label produced after any reasoning trace is parsed as the prediction. If no valid label can be parsed, the prediction is set to \emph{Uncertain}. This baseline measures the performance of the base model under the same three-valued output space as our method.

\paragraph{Qwen Direct SFT.}
The second baseline fine-tunes Qwen3-8B directly on labelled claims, \emph{TFU Training Claims} (see Appendix~\ref{app:main_training_data}). Given a claim, the model is trained to output only the gold label. We use parameter efficient LoRA fine-tuning, applying the language modelling loss only to the label tokens and masking the prompt tokens. This baseline tests whether the proposed argumentative pipeline improves over a model trained directly to map claims to labels using the same supervised data.

\paragraph{Qwen Direct GRPO.}

The third baseline optimises the direct predictor using GRPO with an exact-match classification reward, also using \emph{TFU Training Claims}. For each claim, the model samples label predictions under the same direct classification prompt. A completion receives positive reward if the parsed label matches the gold label and negative reward otherwise. In our implementation, this reward is $+100$ for a correct label and $-100$ for an incorrect or unparseable label. This baseline controls for the possibility that any gains from our full system are due simply to reinforcement learning on the final verification label, rather than to the argumentative structure or the use of formal semantics.

\section{Argument Semantics}
\label{aap:semantics}

\CRchange{In the main experiments} we use the discontinuity-free quantitative argumentation debate semantics, DF-QuAD \citep{Rago2016DiscontinuityFreeDS}. Let
\[
Q=\langle A,R^-,R^+,\tau\rangle
\]
be a QBAF. For an argument $a\in A$, let $\mathrm{Att}(a)=\{b:(b,a)\in R^-\}$ and $\mathrm{Sup}(a)=\{b:(b,a)\in R^+\}$ denote its direct attackers and supporters. DF-QuAD first combines a set of argument strengths using
\[
F(x_1,\ldots,x_k)=
\begin{cases}
0 & \text{if } k=0,\\
1-\prod_{i=1}^{k}(1-x_i) & \text{otherwise.}
\end{cases}
\]
Thus multiple attackers or supporters are aggregated into a single attacking or supporting force. For an argument $a$, let
\[
v^-_a = F\bigl(\{\sigma_Q(b): b\in \mathrm{Att}(a)\}\bigr),
\qquad
v^+_a = F\bigl(\{\sigma_Q(b): b\in \mathrm{Sup}(a)\}\bigr),
\]
where $\sigma_Q(b)$ is the final strength assigned to argument $b$. The final strength of $a$ is then obtained by comparing the aggregate attacking and supporting forces:
\[
\sigma_Q(a)=
\begin{cases}
\tau(a) & \text{if } v^-_a = v^+_a,\\
\tau(a)\bigl(1-(v^-_a-v^+_a)\bigr) & \text{if } v^-_a > v^+_a,\\
\tau(a)+(1-\tau(a))(v^+_a-v^-_a) & \text{if } v^+_a > v^-_a.
\end{cases}
\]
Given fixed base scores and a fixed QBAF, DF-QuAD deterministically returns the final strengths of all arguments.

\section{BSM Full Loss Function}
\label{app:bsm-training}

The full BSM loss is defined over three data resources: the semantic training data
$\mathcal{D}_{\mathrm{sem}}$, the ranking data $\mathcal{D}_{\mathrm{rank}}$, and the consistency data
$\mathcal{D}_{\mathrm{con}}$. It is given by

\[
\begin{aligned}
\mathcal{L}
={}&
\mathbb{E}_{(\hat v_c, v_c) \sim \mathcal{D}_{sem}}
\left[\mathcal{L}_{sem}(\hat v_c, v_c)\right] \\
&+
\lambda_{rank}
\mathbb{E}_{(a_c, \mathcal{P}) \sim \mathcal{D}_{rank}}
\left[\mathcal{L}_{rank}(a_c, \mathcal{P})\right] \\
&+
\lambda_{con}
\mathbb{E}_{(a_c, \mathcal{C}) \sim \mathcal{D}_{con}}
\left[\mathcal{L}_{con}(a_c, \mathcal{C})\right].
\end{aligned}
\]

\section{Ranking and Consistency Experiments}
\label{app:additional-experiments}

The main experiments evaluate BSMs through their effect on the downstream accuracy of the final argumentative verdict. We additionally assess whether the learned base scores satisfy two local desiderata for argument evaluation. First, scores should be stable under paraphrase: semantically equivalent arguments should receive similar base scores. Second, scores should preserve relative argument strength: arguments ranked as stronger should receive higher scores than weaker arguments. To measure this, we conduct experiments on constructed datasets, described in the \emph{Consistency Evaluation Arguments} and \emph{Ranking Evaluation Arguments} sections in Appendix~\ref{app:dataset_eval}. The results are reported in Table~\ref{tab:bsm-diagnostics}. Inconsistency is measured as the variance of the scores assigned to semantically equivalent arguments, averaged across paraphrase groups, so lower values indicate greater stability. Ranking is measured as rank agreement with the expected ordering of argument strengths, so higher values indicate better sensitivity to relative strength.

Both the training regimes using just the priming and semantic loss, as well as those using the auxiliary consistency and ranking losses, substantially improve the consistency of the BSMs compared to the baseline. It is of note that even when not specifically optimising for consistency during training we observe a significant improvement. This is important, as it is the more robust of the two tasks (it is relatively simple for an LLM to generate semantically-equivalent paraphrases of an argument, compared to coming up with an accurate quality ranking), and is a fundamental property of a good base score predictor. The ranking performance worsens on the BSMs only trained on the semantic loss (possibly in part due to noise introduced by the synthetic nature of the dataset), but significantly improves on those trained on the auxiliary losses.

\section{Experimental Details}
\label{apd:experiment-details}

\begin{table}[ht]
\caption{Main performance results (accuracy on TFU and balanced accuracy on AVeriTeC) with 95\% confidence intervals estimated using BCa bootstrap with 10,000 resamples. For the argumentative baselines and our models, A+B indicates which method A was used for argument generation (args) and which method B for generation of base scores. `BSM-Q' stands for `base score model trained on arguments generated by Qwen3-8B', and `BSM-G' stands for `base score model trained on arguments generated by GPT-5'. `sem', `rank' and `con' refer to different losses (see Section~\ref{sec:bsm}). The best results overall are bolded while the best argumentative method results are underlined.}
\centering
\footnotesize
\begin{tabular}{lcc}
\toprule
Variant & TFU & AVeriTeC \\
\midrule
\multicolumn{3}{l}{\textit{Non-Argumentative Baselines}} \\
\midrule
Qwen direct baseline
  & 0.635 (0.599, 0.668) & 0.454 (0.427, 0.483) \\
Qwen direct SFT
  & 0.659 (0.624, 0.692) & 0.446 (0.421, 0.472) \\
Qwen direct GRPO
  & 0.592 (0.556, 0.625) & \textbf{0.485 (0.461, 0.511)} \\
\midrule
\multicolumn{3}{l}{\textit{Argumentative Baselines}} \\
\midrule
Qwen args + constant 0.5 base scores
  & 0.555 (0.519, 0.591) & 0.405 (0.382, 0.427) \\
Qwen args + Qwen base scores
  & 0.585 (0.549, 0.620) & 0.406 (0.381, 0.430) \\
\midrule
\multicolumn{3}{l}{\textit{Argument Generator Model (Ours)}} \\
\midrule
Qwen GRPO args + Qwen base scores
  & 0.591 (0.555, 0.625) & 0.426 (0.401, 0.451) \\
\midrule
\multicolumn{3}{l}{\textit{Base Score Models (Ours)}} \\
\midrule
Qwen args + BSM-Q (primed)
  & 0.583 (0.546, 0.618) & 0.408 (0.383, 0.431) \\
Qwen args + BSM-Q (primed \& sem)
  & 0.588 (0.552, 0.623) & 0.415 (0.388, 0.442) \\
Qwen args + BSM-Q (primed \& sem \& rank \& con)
  & 0.590 (0.553, 0.626) & \underline{0.471 (0.442, 0.501)} \\
Qwen args + BSM-G (primed)
  & 0.588 (0.552, 0.623) & 0.410 (0.386, 0.435) \\
Qwen args + BSM-G (primed \& sem)
  & \textbf{\underline{0.677 (0.643, 0.712)}} & 0.460 (0.431, 0.488) \\
Qwen args + BSM-G (primed \& sem \& rank \& con)
  & 0.584 (0.548, 0.619) & 0.466 (0.438, 0.495) \\
\midrule
\multicolumn{3}{l}{\textit{Combination of Models (Ours)}} \\
\midrule
Qwen GRPO args + BSM-Q (primed \& sem \& rank \& con)
  & 0.581 (0.545, 0.616) & 0.468 (0.439, 0.498) \\
Qwen GRPO args + BSM-G (primed \& sem)
  & 0.668 (0.635, 0.701) & 0.458 (0.429, 0.487) \\
\bottomrule
\end{tabular}

\label{tab:variant-accuracy-cis}
\end{table}

\subsection{\CRchange{Threshold Selection}}
\label{app:thresholds}

\CRchange{We repeat GRPO training with the two selected fixed BSMs and thresholds of $0.4$ and $0.6$. We also optimise symmetric thresholds using training data while keeping the BSM fixed. We consider threshold pairs $(0.5-d, 0.5+d)$, where $d$ is a midpoint between adjacent absolute distances of training claim strengths from $0.5$, and select the pair with the best training performance. For the final two configurations in Table~\ref{tab:threshold-ablation}, thresholds are recalibrated on generated arguments during GRPO. The effects vary across configurations and datasets, indicating that threshold selection can affect the performance of the combined system.}

\begin{table}[tb]
\color{\CRrevisioncolor}
\centering
\footnotesize
\caption{\CRchange{Threshold experiments, reporting accuracy on TFU and balanced accuracy on AVeriTeC. The first two rows use thresholds of $0.4$ and $0.6$. The remaining rows use optimised symmetric thresholds, with recalibration during GRPO for the final two configurations.}}
\label{tab:threshold-ablation}
\begin{tabular}{lcc}
\toprule
Variant & TFU & AVeriTeC \\
\midrule
\multicolumn{3}{l}{\textit{Fixed thresholds at $0.4$ and $0.6$}} \\
\midrule
Qwen GRPO args + BSM-Q (primed \& sem \& rank \& con)
  & 0.538 & 0.485 \\
Qwen GRPO args + BSM-G (primed \& sem)
  & 0.673 & 0.437 \\
\midrule
\multicolumn{3}{l}{\textit{Optimised symmetric thresholds}} \\
\midrule
Qwen args + BSM-Q (primed \& sem)
  & 0.610 & 0.426 \\
Qwen args + BSM-Q (primed \& sem \& rank \& con)
  & 0.560 & 0.475 \\
Qwen args + BSM-G (primed \& sem)
  & 0.678 & 0.461 \\
Qwen args + BSM-G (primed \& sem \& rank \& con)
  & 0.576 & 0.463 \\
Qwen GRPO args + BSM-Q (primed \& sem \& rank \& con)
  & 0.578 & 0.435 \\
Qwen GRPO args + BSM-G (primed \& sem)
  & 0.672 & 0.442 \\
\bottomrule
\end{tabular}
\end{table}

\subsection{\CRchange{Alternative Argumentation Semantics}}
\label{app:qe}

\CRchange{To examine the effect of the choice of argumentation semantics, we retrain and evaluate all four BSM configurations using Quadratic Energy (QE) semantics~\citep{Potyka2018ContinuousDS}, keeping the remaining protocol fixed. Table~\ref{tab:qe-semantics} compares these results with DF-QuAD. Performance varies across configurations and datasets, with neither semantics consistently outperforming the other. These results show that BSM training can also be carried out with QE, while the choice of semantics affects downstream performance.}

\begin{table}[tb]
\color{\CRrevisioncolor}
\centering
\footnotesize
\caption{\CRchange{BSMs retrained with QE, compared with the corresponding DF-QuAD configurations. We report accuracy on TFU and balanced accuracy on AVeriTeC.}}
\label{tab:qe-semantics}
\begin{tabular}{lcccc}
\toprule
& \multicolumn{2}{c}{QE} & \multicolumn{2}{c}{DF-QuAD} \\
\cmidrule(lr){2-3}\cmidrule(lr){4-5}
Variant & TFU & AVeriTeC & TFU & AVeriTeC \\
\midrule
BSM-G (primed \& sem) & 0.681 & 0.445 & 0.677 & 0.460 \\
BSM-G (primed \& sem \& rank \& con) & 0.680 & 0.458 & 0.584 & 0.466 \\
BSM-Q (primed \& sem) & 0.568 & 0.464 & 0.588 & 0.415 \\
BSM-Q (primed \& sem \& rank \& con) & 0.611 & 0.445 & 0.590 & 0.471 \\
\bottomrule
\end{tabular}
\end{table}

\subsection{\CRchange{Experiments with a Second Backbone}}
\label{app:gemma}

\CRchange{We replicate the principal BSM experiments using Gemma-4-E4B-it, with Gemma-generated evaluation arguments and the same DF-QuAD pipeline. Table~\ref{tab:gemma-results} reports the results. The strongest learned scorer on each dataset improves over the strongest training-free argumentative baseline by $2.5$ percentage points on TFU and $3.7$ percentage points on AVeriTeC. The best TFU result approaches direct prediction, while a larger gap remains on AVeriTeC. These results provide evidence that base score learning can benefit a second backbone, although the magnitude of the improvement depends on the model.}

\begin{table}[tb]
\color{\CRrevisioncolor}
\centering
\footnotesize
\caption{\CRchange{Gemma results, reporting accuracy on TFU and balanced accuracy on AVeriTeC. BSM-Gemma and BSM-GPT denote scorers trained on arguments generated by Gemma and GPT, respectively. All argumentative variants use Gemma-generated evaluation arguments.}}
\label{tab:gemma-results}
\begin{tabular}{lcc}
\toprule
Variant & TFU & AVeriTeC \\
\midrule
Gemma direct & 0.524 & 0.444 \\
Gemma args + constant base scores & 0.404 & 0.370 \\
Gemma args + Gemma base scores & 0.492 & 0.364 \\
\midrule
Gemma args + BSM-Gemma (primed \& sem) & 0.517 & 0.404 \\
Gemma args + BSM-Gemma (primed \& sem \& rank \& con) & 0.492 & 0.391 \\
Gemma args + BSM-GPT (primed \& sem) & 0.491 & 0.370 \\
Gemma args + BSM-GPT (primed \& sem \& rank \& con) & 0.509 & \rev{0.374} \\
\bottomrule
\end{tabular}
\end{table}

\subsection{Architecture and Hyperparameters}
\label{app:hyperparams}

\subsubsection{Direct Baselines}

For the direct SFT and direct GRPO baselines, we use LoRA adapters with rank $8$, $\alpha=32$, dropout $0.05$, and target the query, key, value, and output projection matrices. The SFT baseline is trained for three epochs with learning rate $2\times 10^{-4}$, batch size $1$, and gradient accumulation over $8$ steps. The GRPO baseline is trained for three epochs with learning rate $10^{-5}$, batch size $1$, gradient accumulation over $8$ steps, $4$ generations per prompt, temperature $1.0$, and maximum completion lengths of $512$ tokens.

\subsubsection{Argumentative Baselines}

These are the same as in \citep{10.1609/aaai.v39i14.33637}, apart from being restricted to a depth of 1 and not placing a restriction on the width of the frameworks (the original had a variant with depth of 2 and only allowed a maximum of a single supporting and attacking argument in each layer). Furthermore, for this baseline, there is no variation where the claim has an estimated base score (it is always fixed at 0.5), unlike in the original where this is one of the configurations.

\subsubsection{Base Score Model}
During the priming stage of training, we only optimized the BSM regression head, a two-layer MLP with a hidden size of $256$, a ReLU hidden activation and a sigmoid output activation. The model was optimized using an AdamW optimizer with a learning rate of $1\times10^{-4}$, a dropout of $0.1$ and a batch size of $8$. Training was run over a single epoch.

During the second training phase relying on the semantic loss and optionally the ranking and consistency loss, we optimised the regression head as well as LoRA adapters with rank $8$, $\alpha = 32$ and dropout $0.1$ targeting the query, key, value, output, gate, up and down projection matrices. We again used an AdamW optimizer with a learning rate $1\times10^{-4}$, but used gradient accumulation over $8$ steps and gradient checkpointing to reduce memory requirements. In settings where \CRchange{both auxiliary losses} were used, we set $\lambda_\text{rank} = 1$ and $\lambda_\text{con} = 1$ with $m = 0.1$. As for priming, the training was done over a single epoch, with samples for the ranking and consistency losses being randomly sampled at each main training step.

\subsubsection{Reinforcement Learning for Argument Generation}

We fine-tune Qwen3-8B using Group Relative Policy Optimisation (GRPO) with standard HuggingFace Transformers and PEFT. To reduce memory requirements, the model is loaded in bfloat16 precision with gradient checkpointing enabled. We attach LoRA adapters of rank $r = 8$, scaling factor $\alpha = 32$, and dropout $0.1$, targeting the query\rev{, key,} value and \rev{output} matrices. The model is optimised using a learning rate of $1 \times 10^{-5}$ over $2$ epochs, with a per-device batch size of $2$ and gradient accumulation over $4$ steps, yielding an effective batch size of $8$. Each GRPO step samples $2$ completions per prompt, with a maximum prompt length of $512$ tokens and a maximum completion length of $2{,}048$ tokens.

\CRchange{The reward is computed from the final strength of the claim after evaluating its generated supporting and attacking arguments. During GRPO, either a fixed learned BSM or a frozen prompted LLM assigns base scores to the generated arguments, while the claim's own base score remains fixed at $0.5$. Only the argument generator's LoRA parameters are updated. All remaining training hyperparameters are identical across both configurations.}

\subsection{Computational Resources}
\label{app:resources}

\paragraph{Base Score Model} We trained the BSMs on three NVIDIA A100 80GB accessed via a cloud server running a customized distribution of Ubuntu 24.04.2. On this hardware, the generation of the target base scores for priming via direct prompting took approximately $3$ minutes, with the subsequent training on these scores taking approximately $2$ minutes. The second phase of the training took approximately $10$ minutes when not using the ranking and consistency losses and approximately $20$ minutes with these losses.

\paragraph{Argument Generation Model} Similar to the BSMs, we finetuned the argument generators on three NVIDIA A100 80GB GPUs accessed via a cloud server running a customized distribution of Ubuntu 24.04.2. On this hardware, the training for 450 steps took 11 hours using the learned BSM and 19 hours using the LLM-based BSM.

\section{Licences}
\label{app:licenses}

Qwen3-8B \rev{and Gemma-4-E4B-it are} released under Apache-2.0. TruthfulQA and StrategyQA, which underlie TruthfulClaim and StrategyClaim, are released under Apache-2.0 and MIT respectively. The ArgLLMs repository containing the adapted claim-verification datasets is released under the Argumentative LLMs Academic Public Licence, which permits non-commercial academic use. RPB is released under MIT, while AVeriTeC is released under CC BY-NC 4.0. DEBATunE does not appear to provide an explicit redistribution licence, however we reached out to the authors and are still awaiting a response. Other releases by the research group use a BSD 3-Clause License. Synthetic arguments and auxiliary labels generated using GPT-5 and Claude are subject to the applicable OpenAI and Anthropic terms of service.

\section{Limitations}
\label{app:limits}

This work studies claim-only verification without external evidence retrieval. Generated arguments may therefore be plausible but factually unsupported. Incorporating approaches such as RAG into our framework, in the spirit of ArgRAG \citep{ArgRAG-nico-nesy20-25}, is a promising avenue for future work. While ITA provides procedural faithfulness, i.e., the verdict deterministically follows from the considered argumentative structure, scores, thresholds, and semantics, the arguments themselves may be incomplete or incorrect. Our QBAFs are also restricted to depth-one structures, and some supervision signals are synthetically generated, which may introduce artifacts. Finally, we limit ourselves to the task of ternary claim verification. Future work could explore the extension of our framework to domains involving open-ended decision-making, e.g., drawing from the methodology of ArgEval \citep{argeval-2026}, an argumentative approach designed for such domains.

Due to a lack of computational resources, we were limited in the scope of experiments we could carry out. \CRchange{Beyond the targeted threshold studies in Appendix~\ref{app:thresholds}, we were not able to carry out a broader hyperparameter search. The remaining settings} were chosen by estimation and following conventions from the literature rather than empirically validated as the best choice, and therefore our presented results may be an underestimation of the true potential performance of the proposed method.

\section{Safeguards}
\label{app:safeguards}

ITA should be used as an interpretable research tool, not as an automated authority on factual truth. Its outputs should be inspected against trusted evidence, especially in high-stakes domains. Generated arguments and scores should be presented as model-produced rationales, not as verified evidence.

\section{Broader Impacts}
\label{app:impacts}

This work may improve transparency in claim verification by making predictions reconstructible from explicit argumentative structures. However, fluent generated arguments can also make incorrect verdicts appear more credible. Real-world use should therefore pair argumentative explanations with evidence retrieval, source attribution, and human review.



\end{document}